\documentclass[10pt,twocolumn,letterpaper]{article}

\usepackage{cvpr}              %

\usepackage{graphicx}
\usepackage{amsmath}
\usepackage{amssymb}
\usepackage{booktabs}
\usepackage{xspace}
\usepackage{pifont}
\usepackage{tabularx}
\usepackage{paralist}
\usepackage[dvipsnames,table]{xcolor}
\usepackage[subtle]{savetrees}

\usepackage[pagebackref,breaklinks,colorlinks]{hyperref}

\usepackage[capitalize]{cleveref}
\crefname{section}{Sec.}{Secs.}
\Crefname{section}{Section}{Sections}
\Crefname{table}{Table}{Tables}
\crefname{table}{Tab.}{Tabs.}

\def\cvprPaperID{11973} %
\def\confName{CVPR}
\def\confYear{2023}

\definecolor{Gray}{gray}{0.95}
\newcolumntype{a}{>{\columncolor{Gray}}c}

\begin{document}

\newcommand{\gl}{Galactic\xspace}
\newcommand{\hb}{Habitat 2.0\xspace}
\providecommand{\todo}[1]{{\protect\color{red}{\bf [TODO: #1]}}}

\newcommand{\tdy}{Tidy House\xspace}
\newcommand{\reasy}{Rearrange Easy\xspace}

\newcommand{\mobilepick}{\emph{MobilePick \xspace}}
\newcommand{\rearrange}{\emph{Rearrange \xspace}}
\newcommand{\rearrangeheur}{\emph{Rearrange [Safe Drop/Stop] \xspace}}
\newcommand{\rearrangedistr}{\emph{Rearrange [No Distractors] \xspace}}
\newcommand{\rearrangeblind}{\emph{Rearrange [Safe Drop/Stop \& No Vision] \xspace}}

\newcommand{\mobilepicktraintime}{16 minutes}
\newcommand{\spsspeedup}{80$\times$ }

\title{\gl: Scaling End-to-End Reinforcement Learning for Rearrangement\\ at 100k Steps-Per-Second}

\author{
  Vincent-Pierre Berges\thanks{Equal contribution} \\
Meta AI (FAIR)
\and
Andrew Szot\footnotemark[1] \\
Georgia Tech
\and
Devendra Singh Chaplot \\
Meta AI (FAIR)
\and
Aaron Gokaslan \\
Cornell University
\and
Roozbeh Mottaghi \\
Meta AI (FAIR)
\and
Dhruv Batra \\
Meta AI (FAIR), Georgia Tech
\and
Eric Undersander \\
Meta AI (FAIR)
}
\maketitle

\setlength{\abovecaptionskip}{6pt}
\setlength{\belowcaptionskip}{-12pt}

\begin{abstract}

We present \gl, a large-scale simulation and reinforcement-learning (RL) framework for robotic mobile manipulation in indoor environments. 
Specifically, a Fetch robot 
(equipped with a mobile base, 7DoF arm, RGBD camera, egomotion, and onboard sensing) 
is spawned in a home environment and asked to rearrange objects -- by navigating to an object, picking it up, navigating to a target location, and then placing the object at the target location.

\gl is fast. In terms of simulation speed (rendering + physics), \gl achieves 
\textbf{over 421,000 steps-per-second (SPS)} on an 8-GPU node, which is 54x faster than  \hb~\cite{szot2021habitat} (7699 SPS).  
More importantly, \gl was designed to optimize the entire rendering+physics+RL 
interplay since any bottleneck in the interplay slows down training. 
In terms of simulation+RL speed (rendering + physics + inference + learning), 
\gl achieves \textbf{over 108,000 SPS}, which 88x faster than \hb (1243 SPS).  

These massive speed-ups not only drastically cut the wall-clock training time of 
existing experiments, but also unlock an unprecedented scale of new experiments. 
First, \gl can train a mobile pick skill to $>80\% $ accuracy 
in under \mobilepicktraintime, a 100x speedup compared to the over 
24 hours it takes to train the same skill in \hb.
Second, we use \gl to perform the largest-scale experiment to date 
for rearrangement using 5B steps of experience in 46 hours, which is equivalent to 20 years of robot experience.
This scaling results in a single neural network composed of task-agnostic components achieving $85\%$ success in GeometricGoal rearrangement, 
compared to $0\%$ success reported in \hb for the same approach. 
The code is available at  \href{https://github.com/facebookresearch/galactic}{github.com/facebookresearch/galactic} .

\end{abstract}
\vspace{-20pt}

\section{Introduction}

\newcommand{\greencheck}{{\bf \color{OliveGreen}\checkmark}}

\newcommand*\colourcheck[1]{%
  \expandafter\newcommand\csname #1check\endcsname{\textcolor{#1}{\ding{52}}}%
}
\definecolor{bloodred}{HTML}{B00000}
\definecolor{cautionyellow}{HTML}{EED202}
\newcommand*\colourxmark[1]{%
  \expandafter\newcommand\csname #1xmark\endcsname{\textcolor{#1}{\ding{54}}}%
}
\newcommand*\colourcheckodd[1]{%
  \expandafter\newcommand\csname #1checkodd\endcsname{\textcolor{#1}{\ding{51}}}%
}
\colourcheckodd{cautionyellow}
\colourcheck{cautionyellow}
\colourcheck{OliveGreen}
\colourxmark{bloodred}

\begin{table*}[h!]
\resizebox{\textwidth}{!}{
  \newcolumntype{L}{>{\raggedleft\arraybackslash}p{.17\textwidth}} %
 \newcolumntype{R}{>{\raggedright\arraybackslash}p{.19\textwidth}}

    \centering
    \setlength{\tabcolsep}{4pt} %

    \begin{tabularx}{\linewidth}{ p{.281\textwidth}Lrrcccc} %

    \rowcolor{gray!50} \hline

    Arcade RL Sims &  Device & Res  & Sensors & Train SPS & Sim SPS & Photoreal  &  Physics\\
    \hline%
    VizDoom~\cite{Kempka2016ViZDoom,Petrenko2021MegaverseSE} &   1x RTX 3090 & 28×72 & RGB & 18,900 & 38,100 & \bloodredxmark & \cautionyellowcheckodd \\ 
     \rowcolor{gray!50}
        \hline
        Physics-only Sims& Device & Res & Sensors & Train SPS & Sim SPS & Photoreal & Physics \\
        \hline%
        Isaac Gym (Shadow Hand)~\cite{IsaacGym} & 1x A100 & N/A & N/A & 150,000 & -- & \bloodredxmark & \OliveGreencheck \\ %
        Brax (Grasp)~\cite{brax2021github}  &  4x2 TPU v3 & N/A & N/A &  1,000,000 & 10,000,000 & \bloodredxmark & \OliveGreencheck \\ %
        ADPL Humanoid~\cite{APL-PDS}   & 1x TITAN X & N/A & N/A & 40,960 & 144,035 & \bloodredxmark & \OliveGreencheck \\ %
    \hline
    \rowcolor{gray!50}
        EAI Sims & Device & Res. & Sensors & Train SPS & Sim SPS & Photoreal & Physics \\
    \hline %
    iGibson~\cite{Shen2020iGibsonAS,li2021igibson}  & 1x GPU & 128×128 & RGB & -- &  100 & \OliveGreencheck & \OliveGreencheck \\
    AI2-THOR~\cite{deitke2022procthor}  & 8x RTX 2080TI & 224x224 & RGB &  $\approx300$ & 2,860 & \OliveGreencheck & \OliveGreencheck \\ %
    Megaverse~\cite{Petrenko2021MegaverseSE}  & 1x RTX 3090 & 128×72 & RGB &  42,700 & 327,000 & \bloodredxmark & \cautionyellowcheckodd  \\ %
    
       & 8x RTX 2080TI & 128×72 & RGB &  134,000 & 1,148,000 & & \\ 
    LBS~\cite{Shacklett2021LargeBS}  & 1x RTX 3090 & 64×64 & RGB & 13,300 & 33,700  & \OliveGreencheck & \bloodredxmark \\ %
     & 1x Tesla V100 & 64×64 & RGB & 9,000 & -- &  & \\
         & 8x Tesla V100 & 64×64 & RGB & 37,800 & -- &  & \\
    Habitat 2.0~\cite{szot2021habitat,wijmans2022ver}  & 1x RTX 2080 Ti & 128×128 & RGBD & 367 & 1,660 & \OliveGreencheck & \OliveGreencheck\\ %
      & 8x RTX 2080 Ti & 128×128 & RGBD & 1,243 & 7,699 &  & \\ 
    & 1x Tesla V100 & 128×128 & RGBD & 128 & 2,790 &  & \\
         & 8x Tesla V100 & 128×128 & RGBD & 945 & 17,465 &  & \\

     \textbf{Galactic (Ours)}  & 1x Tesla V100 & 128×128 & RGBD  & 14,807 & 54,966 & \OliveGreencheck & \OliveGreencheck \\ %
       & 8x Tesla V100 & 128×128 & RGBD  &  108,806 & 421,194 &  & \\ 
    \hline
    \end{tabularx}
  }
    \caption{{High-level throughput comparison of different simulators. Steps-per-second (SPS) numbers are taken from source publications, and we don't control for all performance-critical variables including scene complexity and policy architecture. Comparisons should focus on orders of magnitude. We show Sim SPS (physics and/or rendering) and training SPS (physics and/or rendering, inference and learning) for various physics-only and Embodied AI simulators. We also describe VizDoom, an arcade simulator which has served as a classic benchmarks for RL algorithms due to its speed. The \cautionyellowcheckodd \xspace for Megaverse and VizDoom represent physics for abstract, non-realistic environments. Among EAI simulators that support realistic environments (photorealism and realistic physics), \gl is \spsspeedup faster than the existing fastest simulator, \hb (108,806 vs 1243 training SPS for 8 GPUs). \gl's training speed is comparable to LBS, Megaverse, and VizDoom, even though LBS doesn't simulate physics and neither Megaverse nor VizDoom support realistic environments. We also compare to GPU-based physics simulators: while these are generally faster than \gl, they entirely omit rendering, which significantly reduces their compute requirements. For \gl, we observe near-linear scaling from 1 to 8 GPUs, with a 7.3x speedup.}}
    \label{tab:sps-comparison-training}
\end{table*}

The scaling hypothesis posits that as general-purpose neural architectures are scaled to larger model sizes and training experience ever increasingly sophisticated intelligent behavior emerges. 
These so-called `scaling laws' appear to be driving many recent advances in AI, leading to massive improvements in computer vision \cite{dalle,rombach2021highresolution,imagenet} and natural language processing \cite{gpt2,gpt3}. 
But what about embodied AI? 
We contend that embodied AI experiments need to be scaled by \emph{several orders of magnitude} to become comparable to the experiment scales of CV and NLP, and likely even further beyond given the multimodal interactive long-horizon nature of embodied AI problems. 

Consider one of the largest-scale experiments in vision-and-language: the CLIP\cite{clip} model was trained on a dataset of 400 million images (and captions) for 32 epochs, giving a total of approximately 12 Billion frames seen during training. 
In contrast, most navigation experiments in embodied AI involve only 100-500M frames of experience \cite{chaplot2020object,Ye2020AuxiliaryTS,maksymets2021thda}. 
The value of large scale training in embodied AI was demonstrated by Wijmans~\etal~\cite{Wijmans2020DDPPOLN} by achieving near-perfect performance on the PointNav task through scaling to 2.5 billion steps of experience.
Since then, there have been several other examples of scaling experiments to 1B steps in navigation tasks~\cite{ramakrishnan2021habitat,yadav2022offline}.
Curiously, as problems become more challenging, going from navigation to mobile manipulation object rearrangement, the scale of experiments have become smaller.
Examples of training scales in rearrangement policies include \cite{szot2021habitat,gu2022multi,wijmans2022ver} training skills in \hb for 100-200M steps and \cite{weihs2021visual} in Visual Room Rearrangement for 75M steps.
In this work, we demonstrate for the first time scaling training to 5 billion frames of experience for rearrangement in visually challenging environments.

Why is large-scale learning in embodied AI hard? 
Unlike in CV or NLP, data in embodied AI is collected through an agent \emph{acting} in environments. 
This data collection process involves policy inference to compute actions, physics to update the world state, rendering to compute agent observations, and reinforcement learning (RL) to learn from the collected data.
These separate systems for rendering, physics, and inference are largely absent in CV and NLP. 

We present \gl, a large-scale simulation+RL framework for robotic mobile manipulation in indoor environments.  
Specifically, we study and simulate the task of GeometricGoal Rearrangement \cite{batra2020rearrangement}, where a Fetch robot~\cite{fetchrobot} equipped with a mobile base, 7DoF arm, RGBD camera, egomotion, and proprioceptive sensing must rearrange objects in the ReplicaCAD~\cite{szot2021habitat} environments by navigating to an object, picking up the object, navigating to a target location, and then placing the object at the target location.

\gl is fast. In terms of simulation speed (rendering + physics), \gl achieves 
\emph{over 421,000 steps-per-second (SPS)} on an 8-GPU node, which is 54x faster than  \hb~\cite{szot2021habitat} (7699 SPS).  
More importantly, \gl was designed to optimize the entire rendering+physics+RL 
interplay since any bottleneck in the interplay slows down training. 
In terms of simulation+RL speed (rendering + physics + inference + learning), 
\gl achieves \emph{over 108,000 SPS}, 
which 88x faster than \hb (1243 SPS).

Our key technical innovations are: (1) integration of CPU-based batch physics with GPU-heavy batch rendering and inference, and (2) a new, approximate kinematic simulation targeted at EAI rearrangement tasks. 
Compared to a ``one simulator, one environment, one process" paradigm, batching yields massive speedups due to memory savings, lower communication overhead, and greater parallelism. 
Meanwhile, we leverage our physics approximations and the reduced complexity of kinematic simulation to reduce our CPU compute and yield further speedups. 

These massive speed-ups not only drastically cut the wall-clock training time of 
existing experiments, but also unlock an unprecedented scale of new experiments. 
First, \gl can train a mobile pick skill to $>80\% $ accuracy 
in under \mobilepicktraintime, a 100x speedup compared to the over 
24 hours it takes to train the same skill in \hb.
Second, we use \gl to perform the largest-scale experiment to date 
for rearrangement using 5B steps of experience in 46 hours, which is equivalent to 20 years of robot experience (assuming 8 actions per second).
This scaling results in a single neural network composed of task-agnostic components (CNNs and LSTMs) achieving $85\%$ success in GeometricGoal rearrangement.
This is impressive performance because (1) the task is extremely long horizon (involving navigation, picking, and placing), (2) the architecture is monolithic and has no mapping modules, task-planning, or motion-planning. 
For context, Habitat 2.0 reported $0\%$ success with a monolithic RL baseline in GeometricGoal rearrangement.
We find the learned policies are able to efficiently navigate, avoid distractor objects, and synchronize base and arm movement for greater efficiency.
Finally, we also show that models trained in Galactic are somewhat robust to zero-shot sim2sim generalization, \ie can achieve $26\%$ success when deployed in 
\hb despite differences in rendering, physics, and underlying controller.

\vspace{-5pt}
\section{Related Work}

\textbf{Scaling Approaches in Embodied AI.} There is a large body of work in speeding up training embodied agents in simulation.
There are three general approaches for increasing efficiency of the overall system: distributing the policy training and inference, increasing the sample efficiency of the learning algorithms, and batch simulation. 

\textit{Distribution and parallelization:} The works in this domain \cite{Espeholt2018IMPALASD,Nair2015MassivelyPM,Wijmans2020DDPPOLN,horgan2018distributed,Stooke2018AcceleratedMF,Menger} achieve efficiency by distributing training across multiple GPUs or nodes and parallelization of the computation. \gl's systems contributions are targeted at rollout computation (inference and environment-stepping) on a single CPU process and single GPU, so it is complementary to many of these approaches \cite{Wijmans2020DDPPOLN,wijmans2022ver,Menger}. %

\textit{Training algorithms:} Various techniques have been developed for the efficiency of training algorithms in interactive settings. Using auxiliary losses \cite{Ye2020AuxiliaryTS,Mirowski2017LearningTN,Yarats2021ImprovingSE,pathakICMl17curiosity,jaderberg2017reinforcement,Shelhamer2017LossII}, offline training \cite{lin2020learning,Munos2016SafeAE,Nachum2018DataEfficientHR,Zhan2020LearningVR,midLevelReps2018,shah2021}, and model-based training \cite{pilco,Gu2016ContinuousDQ,Nagabandi2018NeuralND,Hafner2019LearningLD,hafner2020dreamerv2,Chua2018DeepRL} are some examples that lead to sample efficiency (and typically wall clock time) of the training algorithms. Similarly, since our experiments use a single DD-PPO\cite{Wijmans2020DDPPOLN} policy trained from scratch, \gl can be combined with any of these techniques for further efficiency. 

\textit{Batch simulation:} Batch simulation refers to vectorized physics or rendering to compute updates across multiple environments with one operation in a batched fashion.
This yields large speedups compared to the ``one process, one simulator, one environment" paradigm. 
Our approach is closest to LBS~\cite{Shacklett2021LargeBS} and Megaverse~\cite{Petrenko2021MegaverseSE}, but ours is the first work to combine batch physics and batch rendering to simulate realistic environments and vision sensors.
\cite{Shacklett2021LargeBS} does not support physics, and instead only supports photorealistic, non-interactive scenes and cylinder agents. 
Megaverse~\cite{Petrenko2021MegaverseSE} does not support physics simulation with articulated agents and realistic movable objects, and instead only supports “block worlds” with movable blocks and cylinder agents.

\textbf{Embodied AI Simulators.} There are various Embodied AI simulation platforms \cite{xia2018gibson,Shen2020iGibsonAS,kolve2017ai2,Deitke2020RoboTHORAO,savva2019habitat,szot2021habitat,xiang2020sapien,Gan2020ThreeDWorldAP,Puig2018VirtualHomeSH} for indoor environments that support tasks such as navigation \cite{wortsman2019learning,savva2019habitat}, object manipulation \cite{manipulathor,Srivastava2021BEHAVIORBF}, instruction following \cite{Anderson2018VisionandLanguageNI,shridhar2020alfred}, interactive question answering \cite{gordon2018iqa,Das2018EmbodiedQA}, and object rearrangement \cite{weihs2021visual,szot2021habitat}. The efficiency of the simulators is important for these tasks since they typically involve long task horizons, and the state-of-the-art training algorithms require millions of iterations to converge. \gl supports similar tasks in indoor environments. %

\textbf{Kinematic Simulation.} Some recent work uses "kinematic simulation", in which the robot and objects are moved directly without simulating rigid body dynamics. \cite{szot2021habitat} uses a ``sticky mitten" abstraction for grasping rather than simulating contact physics. It also uses kinematic movement for the base rather than simulating base momentum and wheel forces. \cite{truong2022rethinking} explores training a navigation policy for a quadruped robot in simulation. They show that training with kinematic movement for the base transfers better to real compared to training with full quadruped dynamics. Both of these works use the Habitat 2.0 simulator with underlying CPU-based Bullet physics \cite{pybullet}. For \gl, we introduce a new kinematic simulator optimized specifically for EAI rearrangement tasks.

\section{\gl System}

From an ML systems perspective, RL training can be broken down into rollout computation (collecting experience by interacting with the EAI simulator) and learning (updating the policy). 
Our systems contributions focus on speeding up the experience collection. 
In particular, we optimize batch rollout computation for a single GPU and associated CPU process by (1) integrating CPU-based batch physics with GPU-based batch rendering and DNN inference (\Cref{sec:system_batching}), and (2) introducing a new approximate kinematic simulation optimized for EAI rearrangement tasks (\Cref{sec:system_abstracted_physics}). For our RL experiments, we combine fast rollouts with DD-PPO~\cite{Wijmans2020DDPPOLN} (\Cref{sec:training_setup}), but other approaches to distributed RL are also compatible \cite{wijmans2022ver,Menger}. We discuss throughput and scaling in \Cref{sec:results_benchmarks}.
\gl also supports adding assets from different sources (see \Cref{sec:new_assets} for details).

\subsection{Batching}
\label{sec:system_batching}

Consider non-batched EAI simulators like \cite{szot2021habitat, kolve2017ai2}. They use a ``one Python process, one simulator, one environment" paradigm. The simulator manages a single environment (including physics and rendering) and produces a single set of observations. Additionally, some task-specific logic is implemented directly in Python code, e.g. extracting a robot's pose from the sim and computing reward and other metrics.

Python's cooperative threading model and global interpreter lock make parallelism difficult. To scale this RL training to multiple environments, these approaches typically spawn multiple Python ``env" processes, each hosting its own simulator instance. 
In-memory assets here include CPU data, such as physics collision geometry, and GPU data, such as meshes and textures, and unfortunately they must be duplicated across instances, not shared. Coordinating rollouts and gathering results requires significant interprocess communication, e.g. sending observations between the ``env" and ``main" processes. Observations are batched on the main process and fed to GPU-based batch DNN inference. The number of parallel environments is limited by total system memory, interprocess communication overhead, and other factors and is typically between 16 and 28\cite{szot2021habitat, kolve2017ai2}.

Prior work has shown the benefit of batch simulation over this non-batched paradigm, both for CPU-based physics simulation \cite{Petrenko2021MegaverseSE}, GPU-based physics simulation \cite{IsaacGym, brax2021github,APL-PDS}, and GPU-based rendering \cite{Petrenko2021MegaverseSE, Shacklett2021LargeBS}. For \gl, we batch both CPU-based physics and GPU-based rendering.
For physics, the main Python process hosts a single C++ simulator instance. This instance steps physics for a batch of environments, sharing in-memory CPU assets across environments. Unlike \cite{Shacklett2021LargeBS} which moves nearly all task logic to C++, we retain the flexibility of Python for reward and other task-specific computation. This is implemented as efficient Numpy tensor operations, in contrast to the non-tensor Python code in the non-batched paradigm. We use the Python buffer protocol to ensure we have zero-copy conversion between C++ and Python. 

For rendering, we use Bps3D~\cite{Shacklett2021LargeBS}. A single Bps3D renderer instance renders all environments, sharing GPU memory assets across environments. Scene graph updates are communicated from our physics simulation to the Bps3D renderer efficiently in C++. The renderer outputs a single batch observation per camera sensor, essentially a stack of images from each environment in GPU memory. This is directly consumed by PyTorch GPU-based batch DNN inference, without ever copying pixels to the CPU.
We visualize example observations from \gl in \Cref{sec:task_visuals}.

\begin{figure}[h!]
  \includegraphics[width=\linewidth]{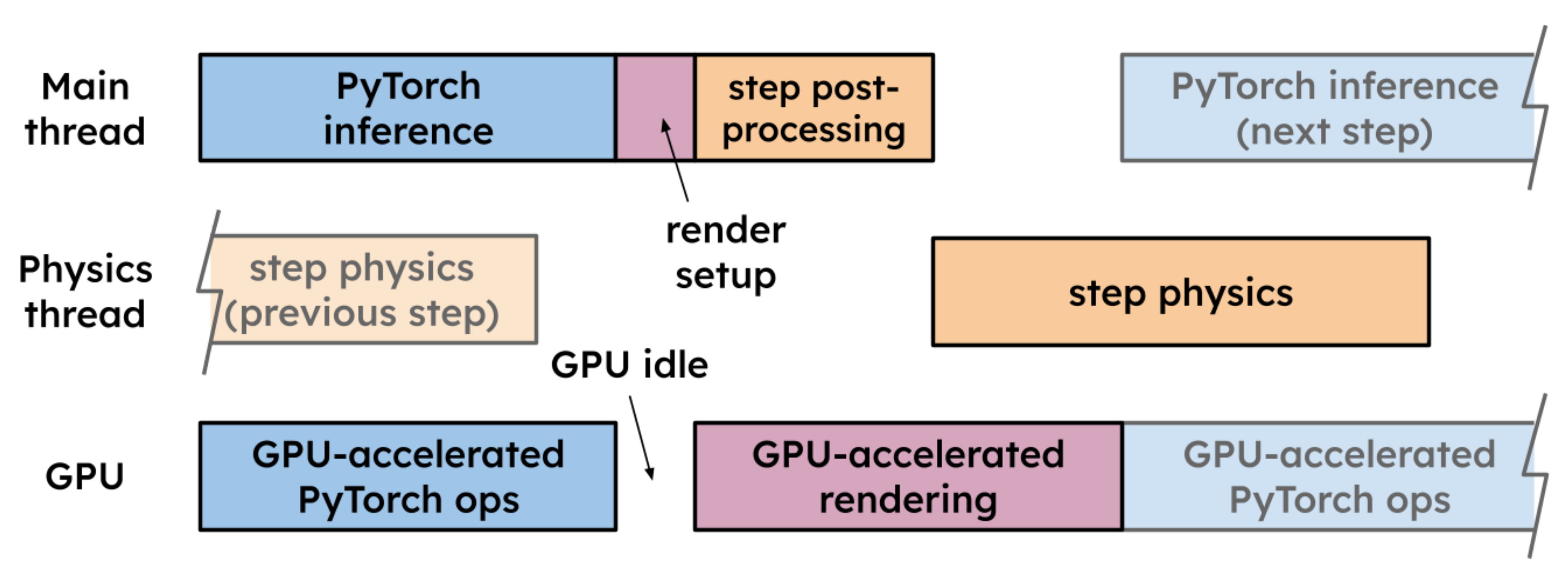}
  \vspace{-20pt}
  \caption{ Rollout timeline for a single batch step, showing our integration of batch physics (orange), batch rendering (purple), and PyTorch DNN inference (blue). CPU-based physics is computed in parallel with GPU-heavy rendering and inference. Physics is fast enough for rollout computation to be primarily GPU-bound.}
  \label{fig:rollout_timeline_figure}
\end{figure}

\Cref{fig:rollout_timeline_figure} shows our integration of batch physics (orange), batch rendering (purple), and PyTorch DNN inference (blue) into the rollout computation loop. ``Step post-processing" refers to reward calculation and other task-specific logic. CPU-based physics is computed in parallel with GPU-heavy rendering and inference. Because of this interleaving, we must accept a one-step-delay: as in \cite{szot2021habitat}, our policy's actions are computed not from the current step's observations, $o_t$, but rather those from the previous step, $o_{t-1}$. Our approximate kinematic sim is fast enough such that we are ``GPU-bound": the bottleneck here is primarily the GPU, with large gaps on both the main and physics threads corresponding to idle CPU. This GPU-bound property is desirable and means \gl will benefit greatly as faster GPUs become available.

Compared to the non-batched paradigm, \gl rollouts have no interprocess communication overhead because all CPU compute happens in a single process. Because of memory savings from batching, our number of environments is larger than the non-batched paradigm (128 versus 16 to 32).

\subsection{Approximate Kinematic Simulation}
\label{sec:system_abstracted_physics}

Batra et al.~\cite{batra2020rearrangement} reviews physics realism for EAI simulators, including agent embodiment and how it interacts with the scene. They describe a spectrum: at the most abstract end are simple cylinder embodiments, ``magic pointer" grasping, and ``virtual backpacks" \cite{savva2019habitat, kolve2017ai2}. At the other end is full rigid-body dynamics simulation: \cite{brax2021github, IsaacGym, APL-PDS}.

\gl's approximate kinematic simulation lies somewhere in the middle. It is kinematic: the robot and objects are moved directly without simulating rigid body dynamics. It uses approximations for detecting and resolving collisions and for simulating object-dropping. Regarding these approximation choices, an important design goal is feasible sim-to-real for policies trained in \gl: our policies should produce detailed, physically-plausible trajectories for an articulated robot and movable objects, requiring only the addition of simple, low-level controllers on real hardware. As a proxy for sim-to-real in this work, we explore sim-to-sim transfer to the Habitat 2.0 dynamic simulator in \Cref{sec:sim2sim}. 

Recent work has employed similar kinematic simulation \cite{szot2021habitat, truong2022rethinking}, but these are built on top of Habitat 2.0 and the existing CPU-based Bullet physics engine~\cite{pybullet}. Meanwhile, we build a new simulator from scratch, leveraging both our physics approximations and the reduced complexity of kinematic simulation to reduce our compute.

\begin{figure*}[h]
  \centering
  \includegraphics[width=0.8\linewidth]{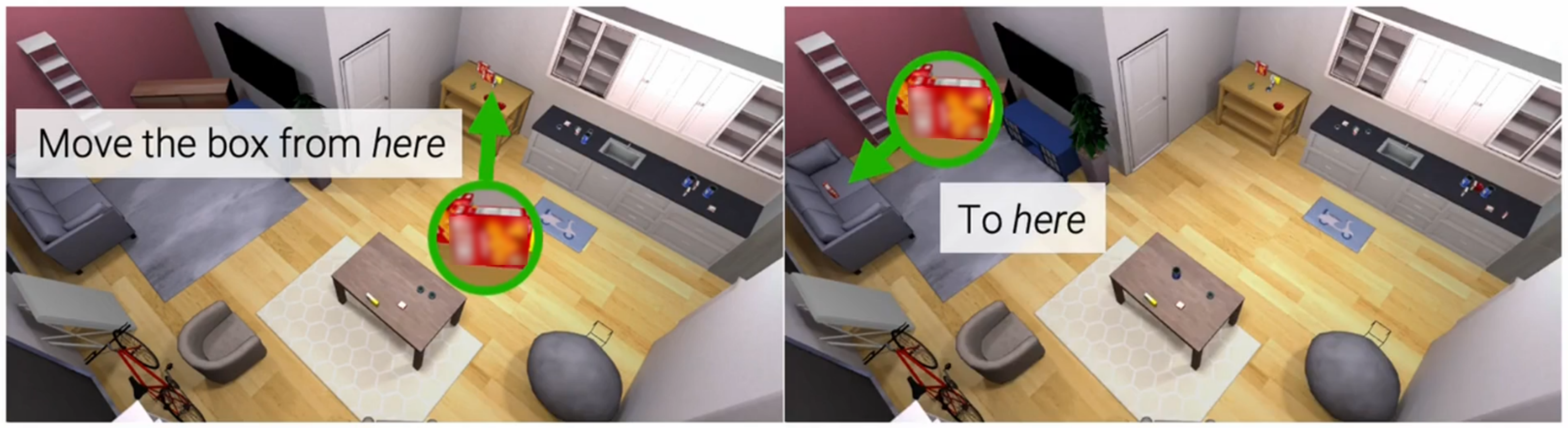}
  \caption{
    Overview of the GeometricGoal rearrangement task. 
    In this task, a Fetch robot must move an object from a start position, specified as a 3D coordinate, to a goal position, also specified as a 3D coordinate, all from egocentric sensing.
    Figure adapted from \cite{habitatrearrangechallenge2022}.
  }
  \label{fig:rearrange_challenge_task}
\end{figure*}

Our action space for the articulated robot includes offsets for base forward/back, base rotation,  and the other degrees of freedom, e.g. arm joint rotation. Inside the simulator, these offsets are applied directly to the robot state, then we use forward kinematics to compute the position of the robot's articulated links. This is kinematic movement: no velocities, forces, or momentum are simulated. This new candidate pose for the robot is tested for overlap (penetration) with the environment. A ``collision" is resolved by either disallowing all robot movement for that step or by sliding the robot base (``allow sliding" is a training hyperparameter). Our sliding implementation is approximate and uses a heuristic search described in \Cref{sec:sliding_heuristics}.

\begin{figure}[h]
  \centering
  \includegraphics[width=7cm]{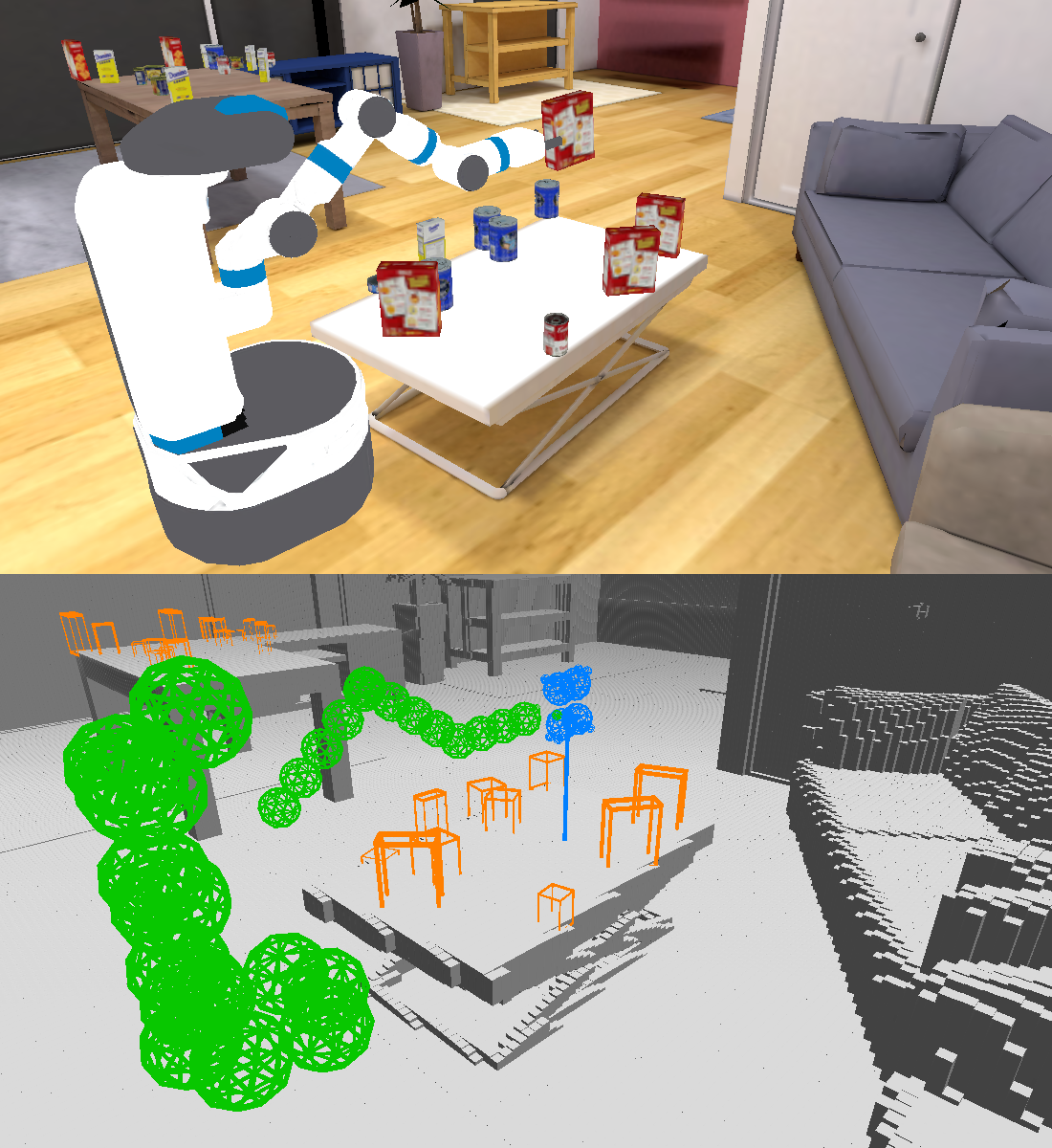}
  \caption{
  Visualization of an environment (top) and its collision geometry (bottom). In the bottom, the robot is represented with green spheres, and the held object with blue spheres. 
  The other movable objects as oriented boxes (orange), and the static environment as a voxel-like ``column grid" data structure (gray).
  }
  \label{fig:gala_collision_geometry}
  \vspace{5pt}
\end{figure}

Collision geometry is generally used in a physics simulator to perform contact and overlap tests. Common primitives include convex hulls and triangle meshes, which can accurately represent real-world shapes using sufficiently high vertex/triangle counts. For our rearrangement task, we assume sub-centimeter tolerances are not important, so we use alternate primitives which are less accurate but faster to query. In particular, we approximate the robot (and the current grasped object, if any) with a set of spheres (green and blue in \Cref{fig:gala_collision_geometry}). We approximate movable objects with oriented bounding boxes (orange). Finally, we approximate the static (non-movable) parts of the environment with a voxel-like structure called a column grid (gray). It has limited precision in the lateral (XZ) direction (3 cm) but full floating-point precision in the vertical (Y) direction, so surface heights are accurately represented. Spheres are authored for the robot manually while bounding boxes and column grids are generated automatically from high-fidelity source meshes in a preprocessing step.

Sphere-versus-environment overlap tests are optimized in a couple ways: (1) we precompute a distinct column grid for each unique sphere radius, so sphere-versus-column-grid tests are simple lookups. (2) Resting movable objects are inserted into a regular-grid acceleration structure. Testing against the entire set of movable objects is fast because we need only iterate over the precomputed set of nearby (50 cm) objects.

Our simulated grasping approximates a real-world suction gripper kinematically without simulating contact or suction forces. Firstly, our action space includes a discrete grasp/release action. In the simulator, if the grasp action is active, we query a small (3 cm) sphere at the tip of the end effector and check for overlap (contact) with movable objects in the scene. If an overlapping object is found, the object is fixed to the end effector. The object moves with the end effector until the release action is performed.

On release, our ``snap-to-surface" operation approximates the entire drop sequence for a released object: falling, landing, and settling. We use a sphere-cast query to find the nearest surface below the object. Instead of allowing movable objects to stack on top of each other, we use an additional heuristic search described in \Cref{sec:placing_heuristics} to find a collision-free resting position nearby. 

We choose to implement our simulator in C++; it runs on the CPU, in contrast to recent GPU-based physics simulators \cite{IsaacGym, brax2021github, APL-PDS}. While these offer greater throughput, CPU-based C++ code has advantages over GPU code: direct access to host memory/disk/network resources, a less restrictive memory model, and greater availability of third-party libraries. In addition, as shown in \Cref{fig:rollout_timeline_figure}, our kinematic sim (``step physics") is faster than GPU-based rendering and PyTorch inference. This makes it ``free" within our GPU-bound rollout computation, as shown by the large idle gaps on the Physics thread. This is in contrast to GPU-based physics simulators, which despite their high throughput, would not be free in this GPU-bound scenario, which would negatively impact overall rollout compute time.

\section{Experiment Setup: Object Rearrangement}
\label{sec:exp_setup} 

\subsection{Task Description}

We study how \gl can accelerate learning in \emph{GeometricGoal object rearrangement}~\cite{batra2020rearrangement} with end-to-end RL. 
We follow the GeometricGoal rearrangement task setup from~\cite{szot2021habitat}, where a robot is tasked with moving an object from a start location to the desired goal entirely from onboard sensing consisting of an RGBD head camera and proprioceptive sensing.
The task is specified via the 3D center of mass for the target object's start location and the 3D center of mass for the object's target location.
To accomplish the task, the agent must navigate through an indoor environment, pick, place, and avoid distractor objects and clutter, all while operating from egocentric sensors.
The agent is a simulated Fetch robot~\cite{fetchrobot}, with a wheeled base, a 7-DoF arm manipulator, a suction gripper, and a head-mounted RGBD camera ($ 128 \times 128$ pixels). 
The episode is successful if the agent places the object within 15cm of the desired position within the episode horizon of 500 time steps and calls the stop action.
We aim to transfer the policies learned in \gl to the \reasy benchmark from the NeurIPS 2022 Habitat Rearrangement Challenge ~\cite{habitatrearrangechallenge2022,szot2021habitat}. 
We implement the same rearrangement task in \gl to facilitate this transfer.

We use an 11-dimension action space in \gl consisting of control for the base, arm, gripper, and episode termination.
2 actions control the linear and angular velocity of the robot base.
7 actions are for delta joint angles for each of the 7 arm joints.
1 binary action is for the grasping and 1 binary action indicates episode termination.

The observation space consists of visual sensors, proprioceptive sensing, and task specification. 
The visual sensor is a $ 128 \times 128$ RGBD, $ 90^{\circ}$ FoV head-mounted camera.
Proprioceptive sensing provides the angles in radians of all 7 joints and if the robot is holding an object. 
From a base egomotion sensor, we derive the relative position of the robot base and end-effector to the object start and goal position as observations.
Additionally, we include an episode step counter in the observation.

We import the datasets and assets from the \reasy benchmark in \hb into \gl.
The \reasy benchmark uses the ReplicaCAD~\cite{szot2021habitat} scene dataset consisting of 105 interactive indoor home spaces.
We load the train dataset of 10k episodes which specify rearrangement episodes in the train split of 63 room layouts in ReplicaCAD. 
We pre-compute the robot's starting position in each episode for greater efficiency. 
Each episode contains one target object to rearrange and 29 distractor objects.

\subsection{Approach: End-to-End RL}
\label{sec:training_setup}

Our approach relies on training a ``sensors-to-actions" policy directly using RL with the task reward alone.
Previous state-of-the-art approaches for mobile manipulation in \hb rely on decomposing the task into the separate skills of navigation, picking, and placing ~\cite{wijmans2022ver,gu2022multi}.
However, such hierarchical approaches suffer from requiring a hand-specified task decomposition and ``hand-off problems" where errors between skills compound.
Our approach avoids these issues by training a single end-to-end policy via RL. 
Previously, such end-to-end approaches achieved little success in the \hb rearrangement tasks~\cite{szot2021habitat}.
However, with the speed of the \gl simulator, we can generate experience fast enough for the end-to-end approach to become a viable approach to rearrangement tasks since it allows gathering the billions of samples needed in only a few days with an 8-GPU compute node.

\noindent\textbf{Reward function.} We define the reward for the rearrangement task largely following the default reward in the \reasy benchmark, but with some modifications for smooth robot motion.
The agent gets a sparse reward for completing the task and picking up the object.
The robot is given a dense reward as the decrease in L2 distance between the end-effector and object as well as between the object and goal.
The robot is given a penalty for large differences in actions at subsequent steps.
Furthermore, to speed up training convergence, we do not allow the robot to call the stop or drop action until the end-effector is within a cylinder of width 0.15m and height 0.3m around the goal position.
We analyze this decision in detail in \Cref{sec:rearrange_analyze}.
Additional details of the rearrangement task and reward function in \gl are in \Cref{sec:add_task_details}.

\noindent\textbf{Architecture and Hyperparameters.} Our primary experiments use a ResNet18~\cite{he2016deep} visual encoder, with a 2-layer 512 hidden unit LSTM, and then separate actor and critic network heads.
We then train this with DD-PPO~\cite{Wijmans2020DDPPOLN}, a distributed version of PPO~\cite{schulman2017proximal}.
We run 128 environments per GPU across 8 GPUs giving 1024 environment instances in total.
We use a policy rollout length of 64, 2 mini-batches per update, and 1 epoch over the rollout data per update.
This setup runs at over 30,000 steps-per-second (SPS), where the policy contains 6 million trainable parameters, 
We provide all training hyperparameters in \Cref{sec:method_details}.

\vspace{-5pt}
\section{Results}

\subsection{Throughput and Scaling}
\label{sec:results_benchmarks} 

In \Cref{tab:sps-comparison-training}, we report sim steps per second (physics and/or rendering) and training SPS (physics and/or rendering, plus inference and learning) for various physics-only and Embodied AI simulators. We also describe VizDoom, an arcade simulator that has served as a classic benchmark for RL algorithms due to its speed.

Among EAI simulators that support realistic environments (photorealism and realistic physics), \gl is \spsspeedup faster than the existing fastest simulator, \hb (108,806 vs 1243 training SPS for 8 GPUs).
\gl's training throughput is comparable to LBS, Megaverse, and VizDoom, even though LBS doesn't simulate physics, and neither Megaverse nor VizDoom supports realistic environments. We also compare to GPU-based physics-only simulators: while these are generally faster than \gl, they entirely omit rendering, which significantly reduces the compute requirements of environment-stepping, inference, and training compared to visual simulators and vision-based policy training. As seen in the last two rows, for our distributed training, we observe near-linear scaling from 1 to 8 GPUs, achieving a 7.3x speedup.

\begin{figure}[!h]
  \includegraphics[width=8cm]{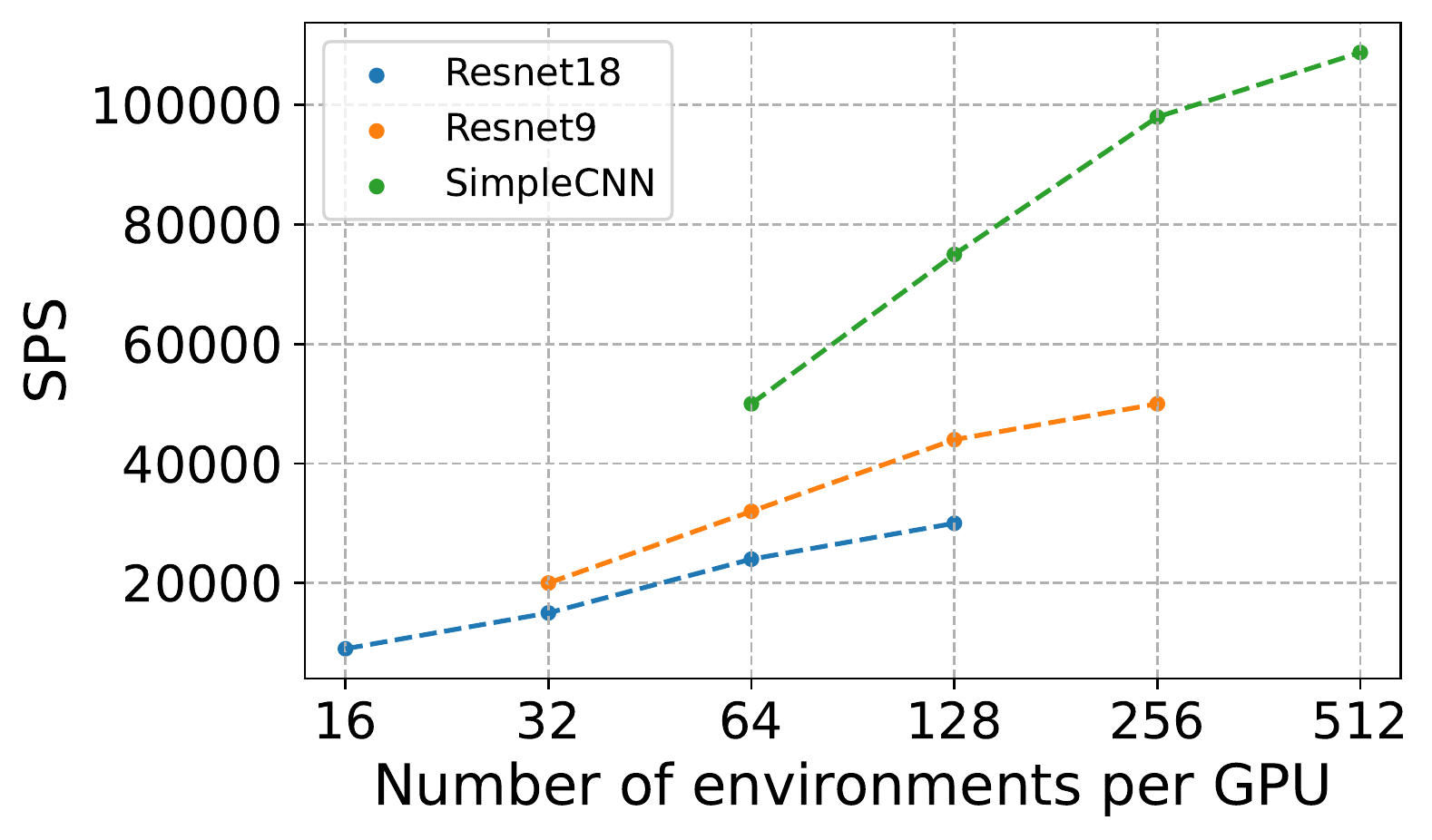}
  \vspace{-5pt}
  \caption{The steps-per-second throughput when varying visual encoder architecture and the number of environments per GPUs, using a single compute node and 8 GPUs. The rightmost point on each curve corresponds to the maximum number of environments we can use before running out of GPU memory. }
  \vspace{5pt}
  \label{fig:gala_sps_curve}
\end{figure}

In \Cref{fig:gala_sps_curve}, we show how the training SPS scales as a function of the number of environments (batch size) and visual encoder on a rearrangement task (described in detail in \Cref{sec:exp_setup}). SPS increases for simpler encoders and larger batch sizes.

\subsection{Sim-to-Sim Results}
\label{sec:sim2sim} 
In this section, we show that we can scale RL training in \gl, achieve a high success rate in \gl, and then zero-shot transfer the policy to \hb.

\noindent\textbf{Training in \gl.} First, we train a policy in \gl using the end-to-end training setup and policy architecture described in \Cref{sec:training_setup}. 
We leverage the fast simulation and easy scaling of \gl to train a neural policy with RL for 5 billion steps of experience.
With 1024 environments across 8 GPUs, training runs at 30,000 SPS which takes 46 hours to train the policy to convergence at 5 billion steps.

\noindent\textbf{Transfer to \hb.} Next, we transfer the policy trained in \gl zero-shot to \hb. 
The physics simulation in \gl is kinematic, while the physics in \hb is dynamic.
In the kinematic simulation of \gl, the actions denote joint delta angles and base velocity, which are set by the simulator on the next step. 
However, in \hb, the arm is controlled via joint motor torques at every step to achieve a target joint state.
Hence, we bridge the dynamics simulation gap by using the policy trained in \gl as a higher-level policy that outputs target joint offset angles in \hb at 10Hz. 
A controller then outputs the joint motor torques to achieve the target joint state set by the policy trained in \gl operating at 120Hz.
The policy takes as input the observations from \hb to output this target delta state.
Observations between \gl and \hb are similar since they use the same scenes and assets.

Other small differences in the simulators affect the sim-to-sim performance.
Since \gl does not simulate dynamics, the dropping mechanism ``snaps" the object to the surface directly below the drop point of the object.
This ignores the effects of the object rolling or bouncing off from the point of contact after the drop.
However, as described in \Cref{sec:training_setup}, the policy is trained to drop the object just above the surface. 
This low drop point minimizes the potential effects of bouncing or rolling due to high drop speeds, helping the sim-to-sim transfer.
Furthermore, collisions are handled differently in the two simulators. In \gl, we ``allow sliding" for training like described in \Cref{sec:system_abstracted_physics}, meaning that the base of the robot will slide when the arm penetrates the environment. However, in \hb, the arm can collide with the surrounding scene.

We compare to training the same policy architecture directly in \hb with RL.
Specifically, we use an identical reward and policy architecture setup as described in \Cref{sec:training_setup} in \hb.  
We train this policy for 50 hours which leads to around 200M steps of training in \hb using the VER trainer~\cite{wijmans2022ver}. 
Training in \hb for 5 billion steps like \gl would take 46 days of training. 
Both the \gl policy and the \hb baseline policy utilize comparable compute resources for training.

\begin{table}[h!]
  \centering
  \resizebox{\columnwidth}{!}{
    \begin{tabular}{cacc}
\toprule
 & \textbf{\gl} & \textbf{\gl $ \rightarrow $ H2.0} & \textbf{H2.0} \\
\midrule
\textbf{Train} &  95.30 {\scriptsize}$ \pm$ 0.67   &  36.70 {\scriptsize}$ \pm$ 0.46   &  0.00 {\scriptsize}$ \pm$ 0.00   \\
\textbf{Eval} &  86.70 {\scriptsize}$ \pm$ 1.06   &  26.40 {\scriptsize}$ \pm$ 0.43   &  0.00 {\scriptsize}$ \pm$ 0.00   \\
\bottomrule
\end{tabular}

  }
  \caption{
    Success rates in the Rearrange task. 
    The middle column shows the zero-shot success of the policy trained in \gl on the fully dynamic \hb (H2.0). 
    The rightmost column is the policy trained purely in H2.0.
    The leftmost column shows success for the \gl trained policy evaluated in \gl (grayed since it reports success in \gl while other columns are in H2.0). 
    Numbers represent mean and standard error across 1k episodes.
  }
  \label{table:sim2sim}
  \vspace{11pt}
\end{table}

\noindent\textbf{Results.} 
We evaluate the success rate of the trained policies on unseen episodes in \Cref{table:sim2sim}.
The top row of \Cref{table:sim2sim} shows the success rate for 1k of the train episodes, and the bottom row shows the success rate over the evaluation dataset of 1k episodes (the ``val" dataset from the \reasy challenge).
The leftmost column (\gl $ \rightarrow $ \gl) of \Cref{table:sim2sim} shows the performance of the policy trained in \gl on the unseen episodes in \gl, demonstrating that the policy can learn generalizable rearrangement behavior from 5B steps of training.
In \hb $ \rightarrow $ \hb, we zero-shot transfer the policy trained in \gl to \hb. 
Despite differences in visuals and dynamics between the simulators, the policy can still achieve $26.40 \% $ success rate without any further training.
This far outperforms training directly in \hb with a similar compute budget as demonstrated by the policy trained directly in \hb achieving no success.
This result is consistent with \cite{szot2021habitat,habitatrearrangechallenge2022} where end-to-end policies also achieve no success.
The policy in \hb learns too slow, and even after 200M steps, it only learns to pick the object up but struggles to place it.
In \Cref{sec:more-hb-results}, we further analyze policy in \hb and the errors in the zero-shot transfer.

\subsection{Analyzing Rearrangement Settings in \gl}
\label{sec:rearrange_analyze} 

In this section, we analyze what properties of the rearrangement task structure are important for RL training.
\gl allows us to answer these questions at scale by training policies fully to convergence, even if it takes billions of steps.

\begin{figure}[h]
  \includegraphics[width=8cm]{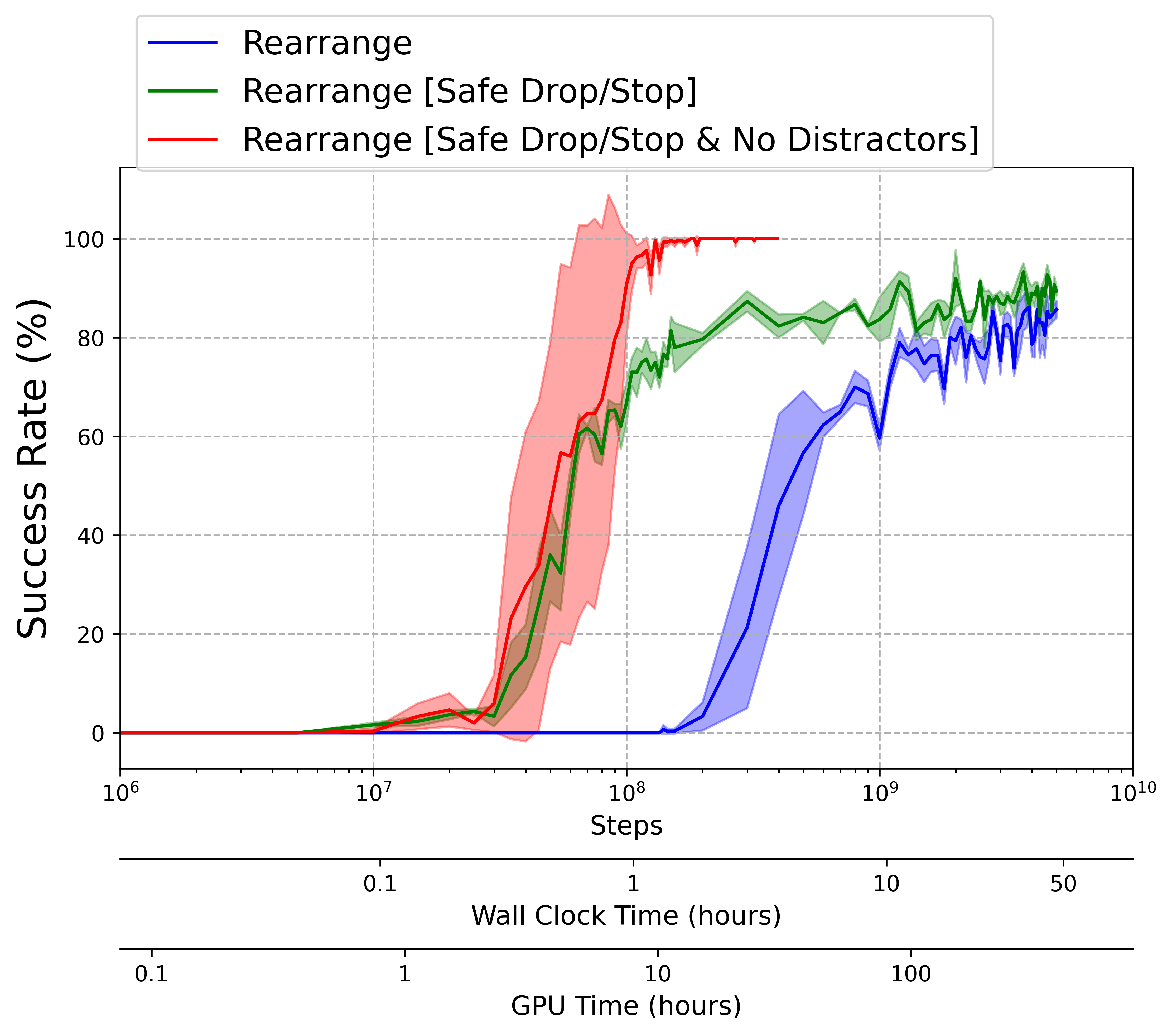}
  \vspace{-7pt}
  \caption{
    Evaluation success rate of different \rearrange tasks on 100 held-out episodes with mean and std across 3 seeds. 
  }
  \vspace{10pt}
  \label{fig:3exp_figure}
\end{figure}

We analyze three variations of the rearrange task with different conditions around the termination and drop action.
\begin{compactitem}
 \item   \rearrange: The ``default" version of rearrangement where the policy may drop the object at any point or call the stop action at any point in the episode. 
  \item \rearrangeheur: Same as \rearrange except the policy has a heuristic condition around when the stop and drop actions are executed.
    The drop action is ignored except when the robot's end-effector is within a cylinder of radius 0.15m and height of 0.3m around the goal location.
    The stop action is ignored before the robot has dropped the object.
  \item \rearrangedistr: A version of \emph{\rearrangeheur} where there are no other objects in the scene other than the object the robot needs to move.
\end{compactitem}
Policies are trained for 5 billion steps in the same training setup as \Cref{sec:sim2sim}. 
\Cref{fig:3exp_figure} compares the learning curves for each setting where each point on the curve is the policy checkpoint evaluated for 100 validation episodes (unseen configurations of objects and scenes) from the \tdy dataset.
When evaluating the last checkpoint on 1000 evaluation episodes, we find that \rearrangeheur achieves 84.73\% success rate while \rearrange achieves 79.63\% success rate, despite using the same sensor inputs.
This demonstrates that the challenges of learning the stop and drop actions harm learning.
Furthermore, the gap between \rearrangeheur with a success rate of 84.73\% and \rearrangedistr with a 99.5\% success rate after only 500 million steps shows that the presence of distractor objects makes learning more difficult.

\subsection{Speeding Up Mobile Pick}

We also show that \gl can rapidly train policies in an additional mobile pick task.
In the mobile pick task, the robot is spawned within a 2-meter radius of the object to pick up and is provided the start coordinates of the object.
The agent then controls the base and arm to navigate to and then pick up the object.
The episode is successful if the agent picks up the correct object within the episode horizon of 300 steps.
We use the policy architecture described in \Cref{sec:training_setup} with a ResNet18 visual encoder for policies in \gl and \hb.

\begin{figure}[h]
  \centering
  \includegraphics[width=7cm]{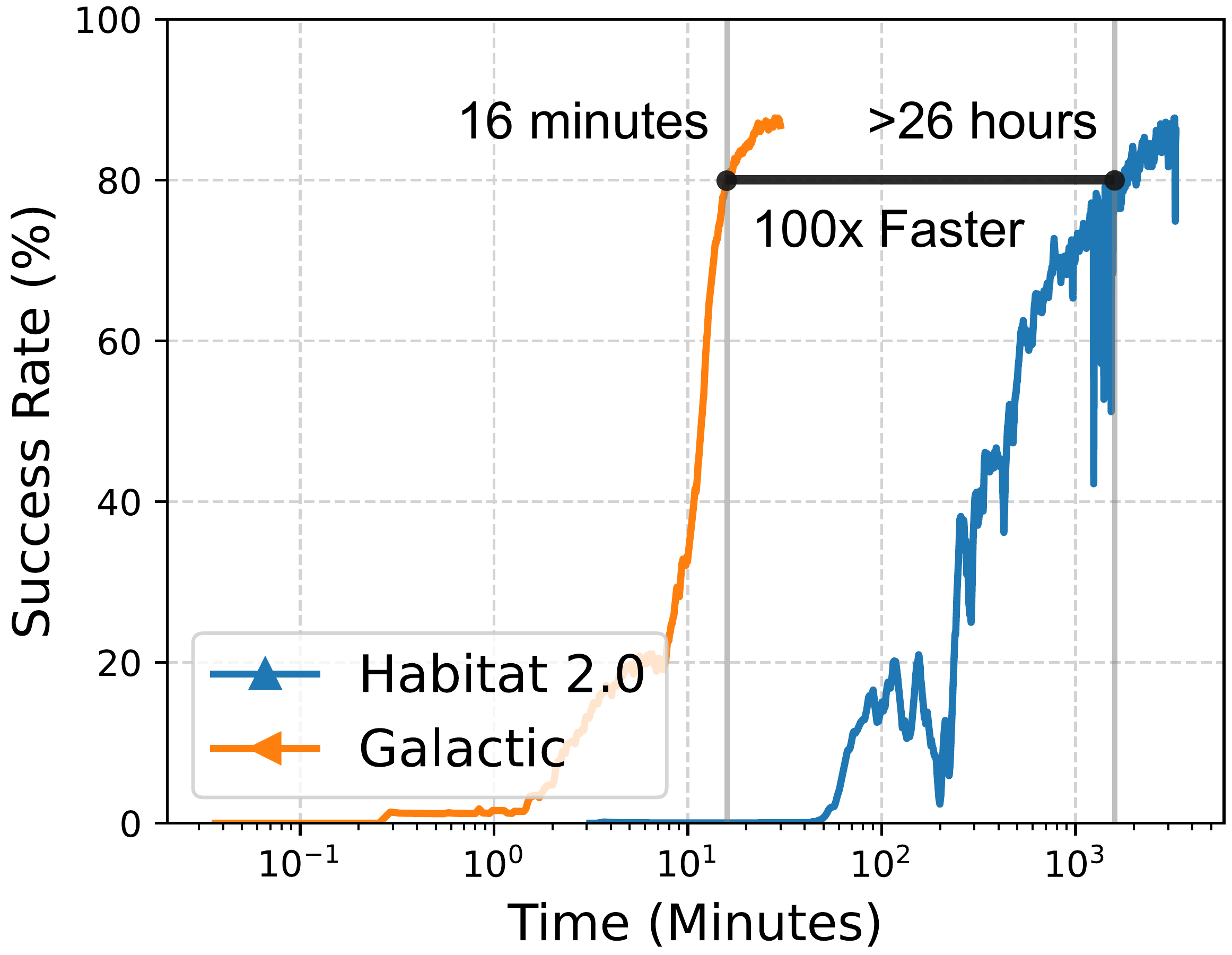}
  \vspace{-4pt}
  \caption{
    Wall-clock time versus success rate in the mobile pick task between training policies in \gl and \hb. 
    \gl takes \mobilepicktraintime to train a policy to 80\% accuracy while \hb requires over 26 hours to train a policy to 80\% accuracy. 
  }
  \label{mobilepick_gl}
\end{figure}

The plot in \Cref{mobilepick_gl} shows that \gl is capable of training policies in \gl 100$\times$ faster than in \hb.
To reach an 80\% success rate, training in \gl requires \mobilepicktraintime of wall-clock time, while reaching the same success in \hb requires over 26 hours. %
This 100$\times$ speed-up validates that \gl's higher SPS translates directly to faster wall-clock time-to-convergence. It also demonstrates that \gl can be used to accelerate even tasks less complex than full rearrangement.

\section{Conclusion}
We propose \gl, a framework for rendering, physics, and RL training for embodied tasks at a massive scale. To achieve high throughput for experience collection, we develop an approximate kinematic simulator optimized for embodied rearrangement tasks and propose a batch processing approach to jointly render observations and simulate the physics of the world. 
\gl processes more than 100,000 steps per second in RL training. 
Learning a mobile picking skill takes less than \mobilepicktraintime, while training for the same task in \hb, one of the fastest embodied simulation frameworks, requires more than 26 hours. 
We also show that policies trained in \gl generalize to zero-shot to \hb. We hope that \gl opens up new avenues in embodied AI research by enabling large-scale training.

\noindent \textbf{Acknowledgements.}
The Georgia Tech effort was supported in part by NSF, ONR YIP, and ARO PECASE. The views and conclusions contained herein are those of the authors and should not be interpreted as necessarily representing the official policies or endorsements, either expressed or implied, of the U.S. Government, or any sponsor. We thank Brennan Shacklett for his help with Bps3D and systems advice. We also thank Erik Wijamns for his help with the training pipeline and general advice on RL training.

{\small
\bibliographystyle{ieee_fullname}
\bibliography{egbib}
}

\clearpage
\appendix

\section{Additional Task Details}
\label{sec:add_task_details} 

The full reward function of the \rearrange task is described in \Cref{8692}.
\begin{equation}
\label{8692}
   r_t = 10\mathbb{I}_{success} + 5\mathbb{I}_{pick} + \Delta^{obj}_{arm} + \Delta^{goal}_{obj} - 0.001 C_t
\end{equation}

Where: 
\begin{itemize}
  \item $\mathbb{I}_{success}$ is the indicator for task success.
  \item $\mathbb{I}_{pick}$ is the indicator if the agent just picked up the object.
  \item $\Delta^{obj}_{arm}$ is the change in Euclidean distance between the end-effector ($arm$) and the target object ($obj$). If $d_t$ is the distance between the two at timestep $t$, then $\Delta^{obj}_{arm} = d_{t-1}-d_t)$.
  \item $\Delta^{goal}_{obj}$ is the change in Euclidean distance between the object (obj) and the goal position (goal). 
  \item $C_t$ Is the squared difference in joint action values between the current and previous time step. If $a^k_t$ is the action for timestep t for moving joint $k$ then $C_t = \sum_k{(a^k_t - a^k_{t-1})^2}$%
\end{itemize}
The reward signal for the \emph{mobile pick} is identical, but the task ends in a success if the robot picks the correct object ($\mathbb{I}_{success} = \mathbb{I}_{pick}$).

The action space of the monolithic policy consist of 11 actions:
\begin{itemize}
    \item 7 continuous actions controlling the change to the joints angles. These actions are normalized between $-1$ and $1$ with the minimum and maximum value corresponding to the maximum change to the joint angles allowed in each direction per step.
    \item 1 continuous action between $-1$ and $1$ corresponding to the robot moving forward. An action with value of $1$ corresponds to the robot moving forward by $10$cm and $-1$ to the robot moving backward by $10$cm in the simulation.
    \item 1 continuous action between $-1$ and $1$ corresponding to the robot rotating. An action with value of $1$ corresponds to the robot rotating in clockwise by $5^\circ$ and $-1$ to the robot rotating counter-clockwise by $5^\circ$.
    \item 1 discrete action with 2 options corresponding to the robot attempting to grasp or release an object. If the value is $0$, and the robot is holding an object, the robot will attempt to release it. If the value is $1$ and the robot is not holding an object, the robot will attempt to grasp.
    \item 1 discrete action with 2 options corresponding to the robot attempting to terminate the episode. If the value is $0$ the robot will continue the task. If the value is $1$, the robot will signal that the task is completed.
\end{itemize}

\section{Method Details}
\label{sec:method_details} 

More details about the method architecture here.

Our Hyperparameters are described in Table  \ref{hyperparams_table}.
\begin{table}[h!]
\centering
\begin{tabular}{||c|c||} 
 \hline
 Hyperparameter & Value \\ [0.5ex] 
 \hline\hline
 start learning rate & $3.5 \times  10^{-4}$  \\ 
  \hline
   end learning rate & $0$  \\ 
  \hline
     learning rate schedule & $linear$  \\ 
  \hline
     entropy coefficient & $1 \times  10^{-3}$  \\ 
  \hline
       clip gradient norm & $2.0$  \\ 
  \hline
       time horizon & $64$  \\ 
  \hline
       number of epochs per updates & $1$  \\ 
  \hline
       number of mini batches per updates & $2$  \\ 
  \hline
       RGB and Depth image resolution & $128 \times 128$  \\ 
  \hline
 image encoder & $ResNet18$  \\ 
  \hline
normalized advantage & $true$  \\ 
  \hline
\end{tabular}
\caption{Hyperparameters used for DD-PPO training in \gl}
\label{hyperparams_table}
\end{table}

To calculate the entropy of this action space for entropy regularization in DD-PPO, we add the entropy of the discrete and continuous actions distributions together without any scaling.

The SimpleCNN model we use consists of 3 convolution layers followed by a fully connected layer. The kernel sizes for the three convolution layers are $8 \times 8$, $4 \times 4$ and $3 \times 3$, the strides are $4 \times 4$, $2 \times 2$ and $1 \times 1$ and there is no dilation nor padding. This is the same SimpleCNN visual encoder used in \hb. The size of the models used are described in Table \ref{model_sizes}.

\begin{table}[h!]
\centering
\begin{tabular}{||c|c||} 
 \hline
 Model & Total number of parameters \\ [0.5ex] 
 \hline\hline
 SimpleCNN & $4,046,999$  \\ 
  \hline
 ResNet9 & $4,338,007$  \\ 
  \hline
 ResNet18 & $5,906,647$  \\ 
  \hline
\end{tabular}
\caption{Model sizes for the different visual encoders used. This includes the visual encoder, the actor, and the critic.}
\label{model_sizes}
\end{table}

\section{Further \hb Results}
\label{sec:more-hb-results}

\begin{figure}[!h]
  \centering
  \begin{subfigure}[t]{0.95\columnwidth}
    \includegraphics[width=\textwidth]{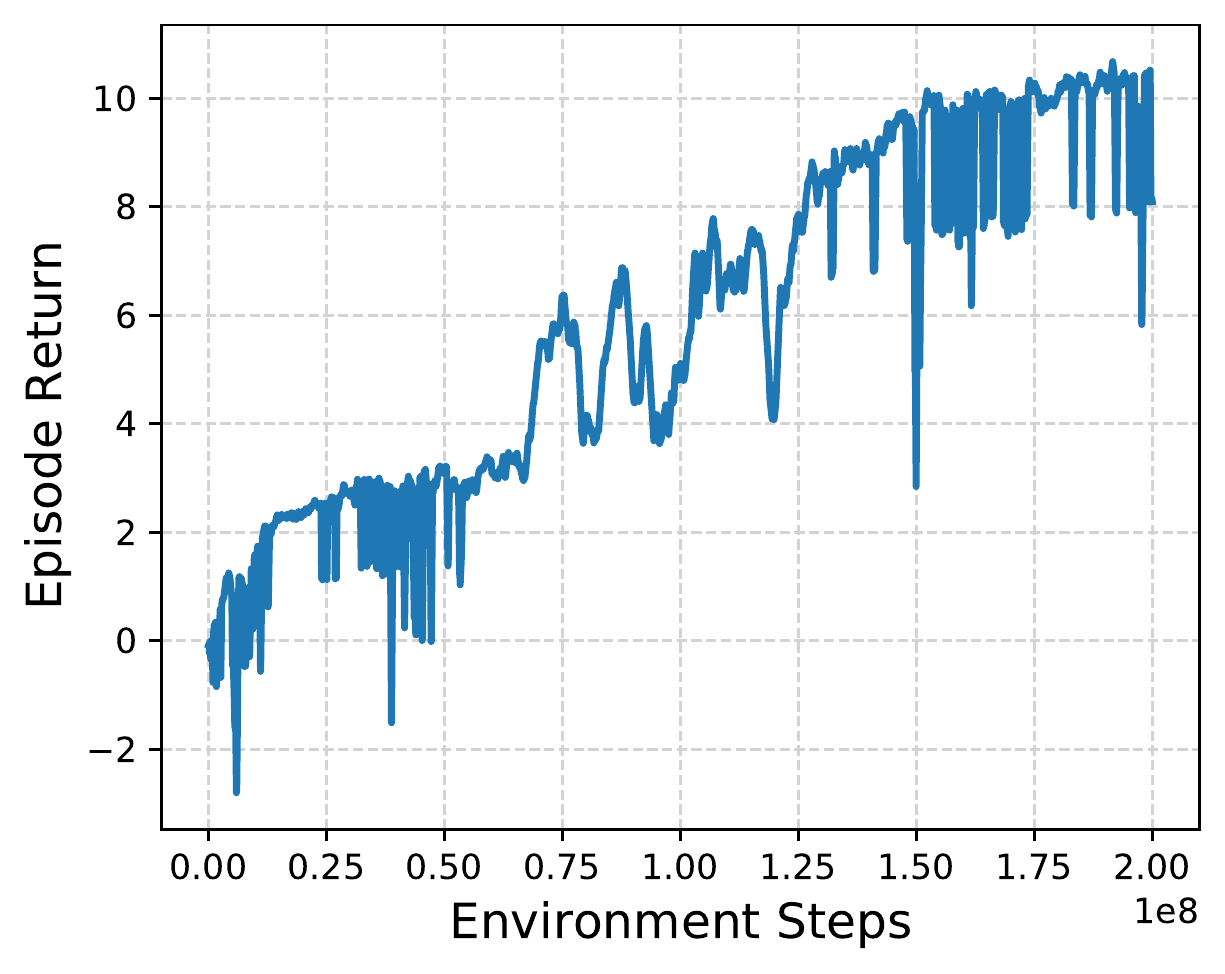}
    \caption{Reward}
    \vspace{10pt}
    \label{fig:h2:reward}
  \end{subfigure}
  \begin{subfigure}[t]{0.95\columnwidth}
    \includegraphics[width=\textwidth]{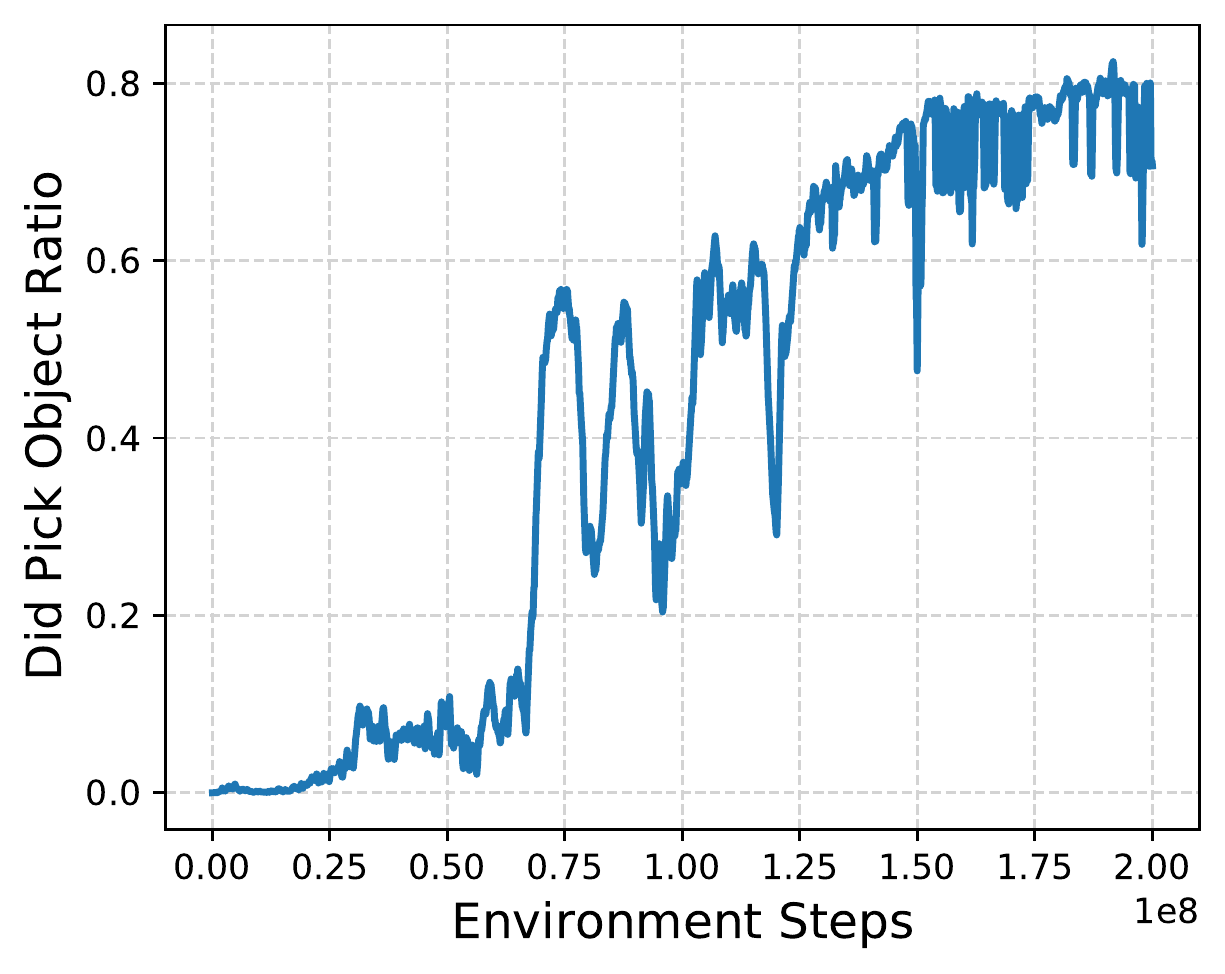}
    \caption{Picked object ratio}
    \vspace{10pt}
    \label{fig:h2:pick_obj}
  \end{subfigure}
  \caption{
    Learning curves for the policy trained purely in \hb. \Cref{fig:h2:reward} shows the episode reward does not saturate even after 200M training steps.
    \Cref{fig:h2:pick_obj} shows that even though the agent is never successful, it still learns to pick the object. 
  }
  \label{fig:h2_learn}
\end{figure}

First, we analyze the poor performance of the policy purely trained in \hb, which achieves no success in the \Cref{table:sim2sim}.
\Cref{fig:h2:reward} shows the reward learning curve during training.
This learning curve demonstrates that even after 200M steps of training, the reward is still increasing, which provides evidence for the necessity of \gl to scale training.
In this training time, the agent reliably learns to pick the object around $80\%$ of the time as shown by the training plot in \Cref{fig:h2:pick_obj} comparing the fraction of the time the robot picked the object within an episode versus the number of training steps.

Next, we analyze the source of errors in the zero-shot transfer from \gl to \hb.
We show the drop in performance is not due to the dynamic arm control by comparing to transferring to \hb with a kinematic arm controller instead of a dynamics-based torque controller. 
The agent with the kinematic arm controller achieves a $ 29.7\% $ success rate on the ``Eval" dataset, barely any better than the $ 26.4\% $ success rate the dynamics-based torque controller achieves.

\section{Additional Task visuals}
\label{sec:task_visuals} 
\begin{figure}[h!]
  \includegraphics[width=8cm]{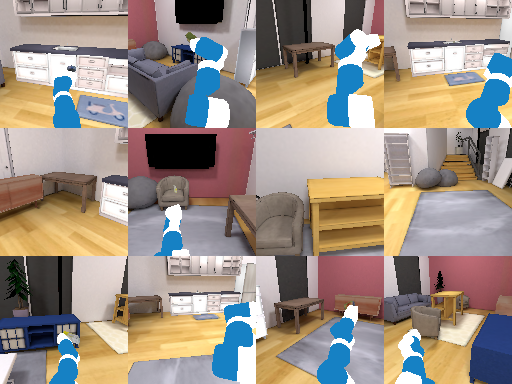}
  \caption{Samples of RGB observations collected in \gl.}
  \label{gala_rgb}
\end{figure}

\begin{figure}[h!]
  \includegraphics[width=8cm]{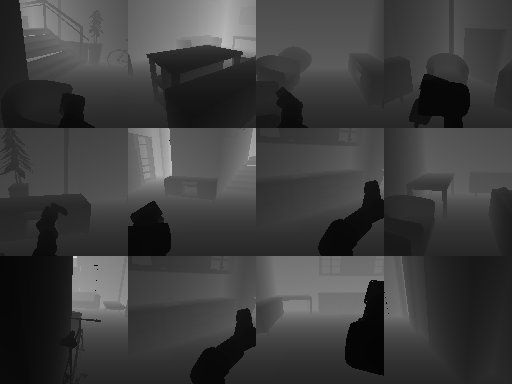}
  \caption{Samples of Depth observations collected in \gl.}
  \label{gala_depth}
\end{figure}

\begin{figure}[h!]
  \includegraphics[width=8cm]{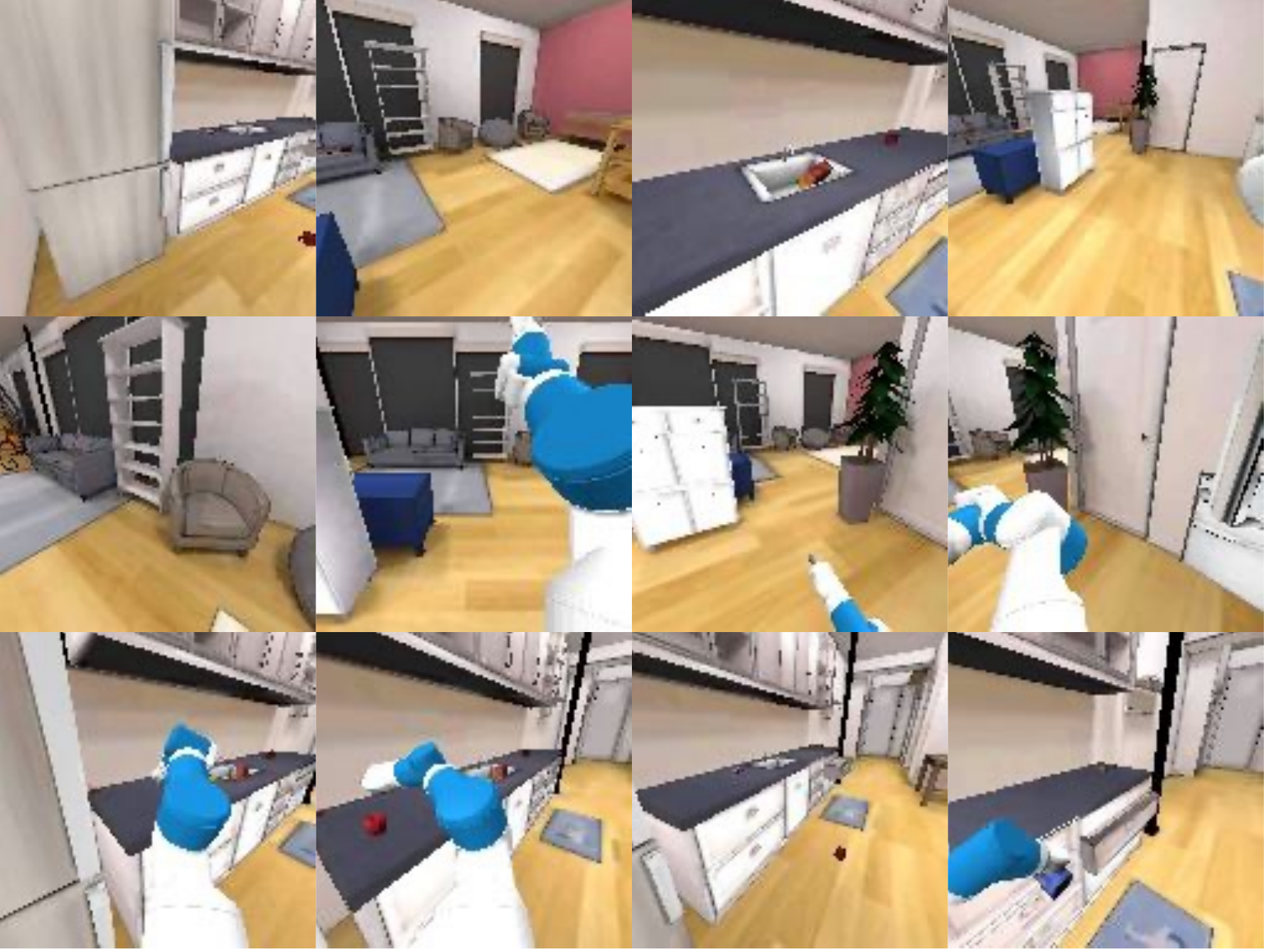}
  \caption{Samples of RGB observations collected in \hb.}
  \label{hb_rgb}
\end{figure}

\begin{figure}[h!]
  \includegraphics[width=8cm]{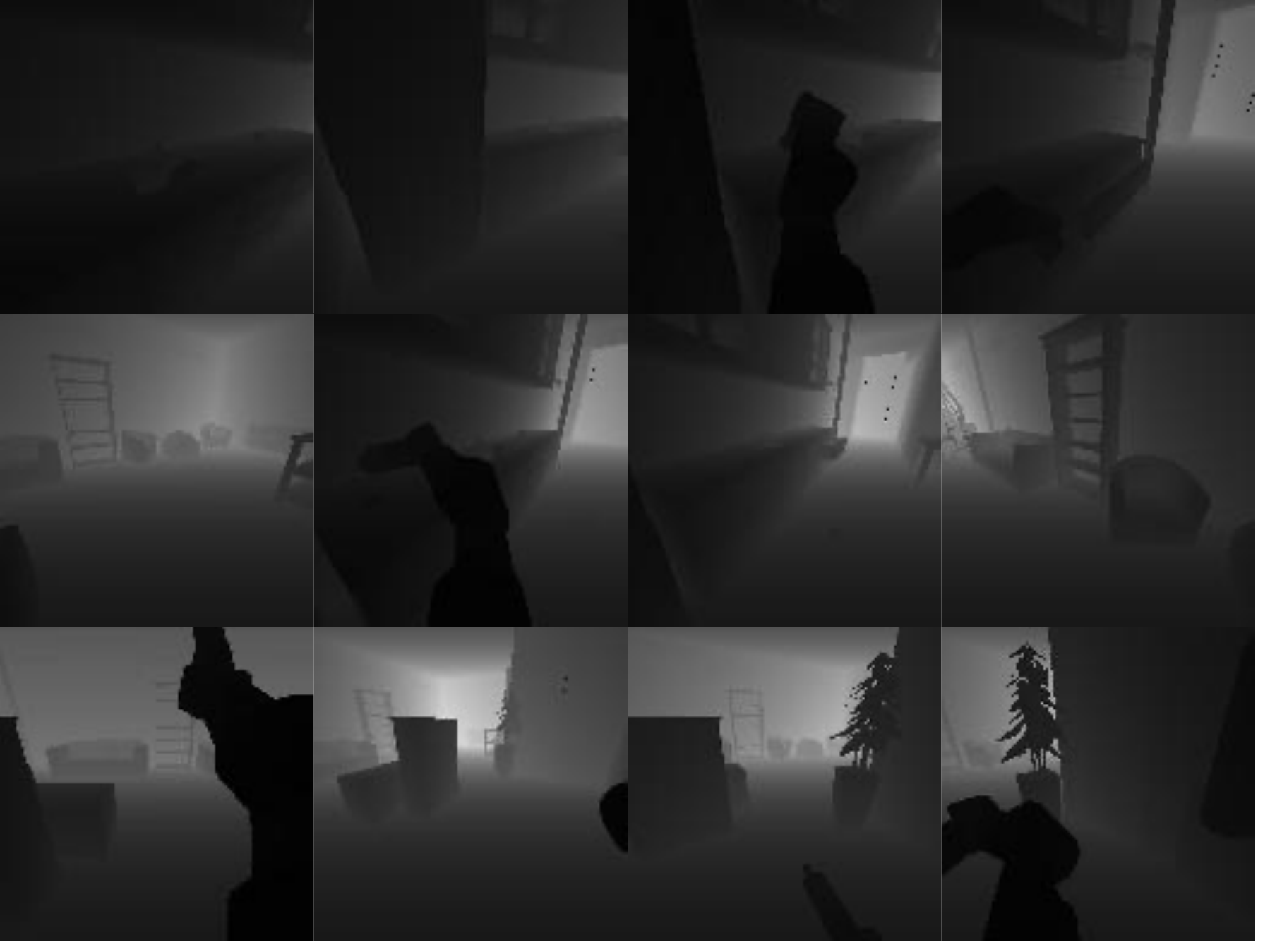}
  \caption{Samples of Depth observations collected in \hb.}
  \label{hb_depth}
\end{figure}

In this section, we visualize observations rendered using the \gl simulator. \Cref{gala_rgb} are examples of 128 $\times$ 128 RGB images used for training. \Cref{gala_depth} are examples of 128 $\times$ 128 depth images used for training. We also visualize observations rendered using the \hb simulator also at $ 128 \times 128$ in \Cref{hb_rgb} and \Cref{hb_depth}.


\section{Training for 15 Billion steps}

To show the usefulness of training for several billions of steps, we trained the \rearrange task defined in \Cref{sec:exp_setup} for 15 billion steps. Training and validation success rates are still improving, showing that training still hasn't converged, even after 15 billion steps.

\begin{figure}[h!]
  \includegraphics[width=8cm]{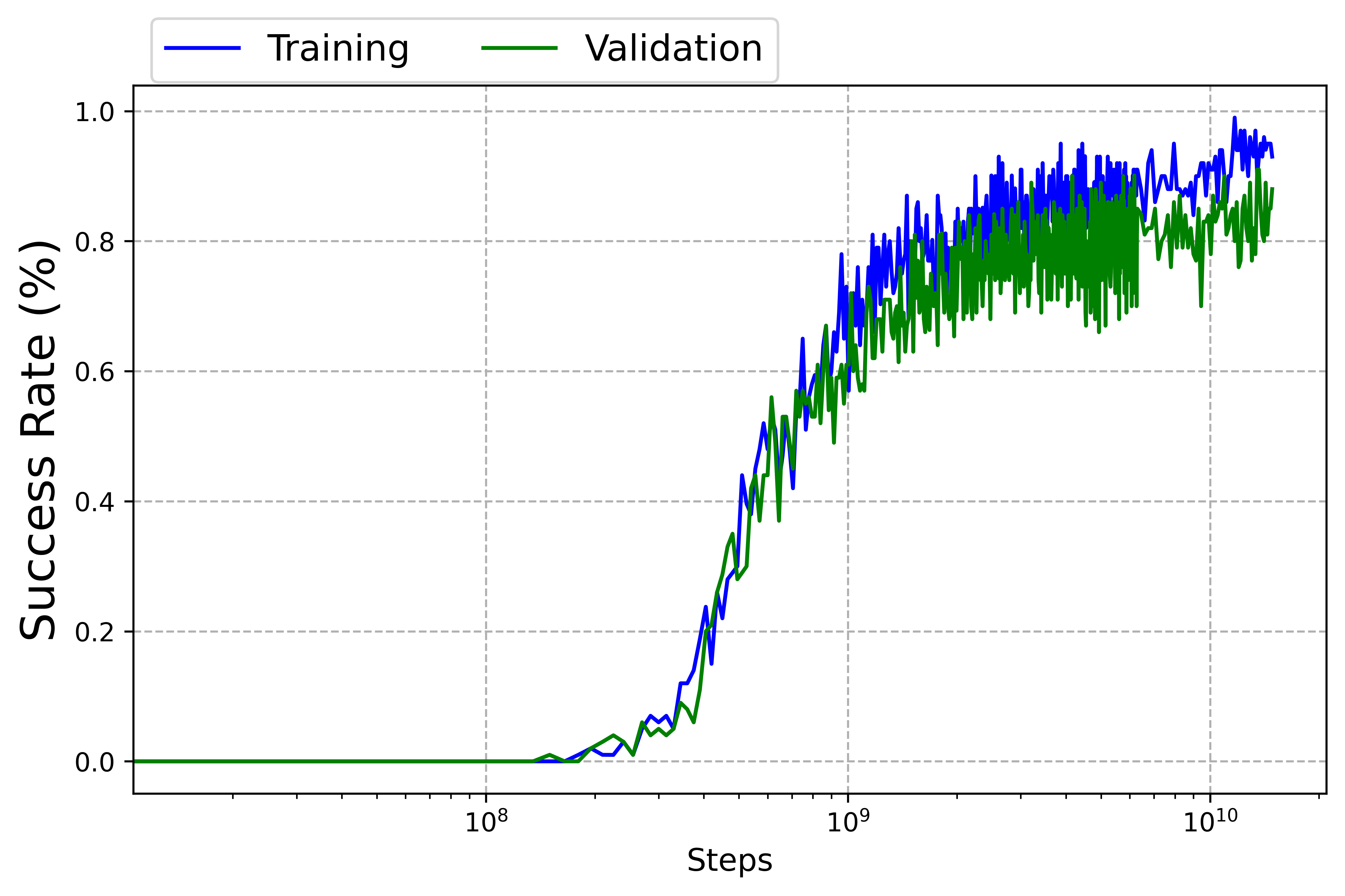}
  \caption{Training and evaluation curves for a 15 billion steps training run. Each checkpoint is evaluated on 100 training or validation episodes.}
  \label{gala_15b}
\end{figure}

\section{Performance Timings}
\label{sec:add_perf_timings} 

\begin{table}[h!]
\centering
\begin{tabular}{ ||p{9em}||p{4em}|p{5em}|| } 
  \hline
  Visual Encoder & ResNet18 & SimpleCNN \\
  \hline
  Number of Envs & 128 & 512 \\
  \hline
  \hline
  PyTorch Inference & 7.52 & 8.80 \\
  \hline
  Render Setup & 0.77 & 1.47 \\
  \hline
  Step Post-processing* & 1.84 & 2.05 \\
  \hline
  Step Physics* & 1.78 & 3.76 \\
  \hline
  GPU Rendering & 2.27 & 9.68 \\
  \hline
  Additional CPU & 0.74 & 1.00 \\
  \hline
  \hline
  Total & 11.30 ms & 20.95 ms \\
  \hline
\end{tabular}
\caption{Timing breakdown of a single batch rollout step, for two configurations in milliseconds. * Post-processing and Physics are interleaved with GPU rendering and PyTorch inference and don't contribute to total rollout step time. See also \Cref{fig:rollout_timeline_figure}. 1x Tesla V100, 10x Intel Xeon Gold 6230 CPU @ 2.10GHz, 128x128 RGBD sensors. }
\label{perf_timings_rollout_table}
\end{table}

\begin{table}[h!]
\centering
\begin{tabular}{ ||p{9em}||p{4em}|p{5em}|| } 
  \hline
Visual Encoder & ResNet18 & SimpleCNN \\
  \hline
Number of Envs & 128 & 512 \\
  \hline
  \hline
Compute Rollouts & 726 & 1289 \\
  \hline
Update Agent & 1381 & 856 \\
  \hline
  \hline
Total & 2204 ms & 2215 ms \\
  \hline
  \hline
Training SPS & 3716 SPS & 14791 SPS \\
  \hline
\end{tabular}
\caption{Timing breakdown of a single train update for two configurations in milliseconds. 1x Tesla V100, 10x Intel Xeon Gold 6230 CPU @ 2.10GHz, 128x128 RGBD sensors, 64 batch rollout steps. }
\label{perf_timings_train_update_table}
\end{table}

\section{Additional Collision-Detection Details}
\label{sec:add_collision_detection_details} 

In this section, we'll expand on \Cref{sec:system_abstracted_physics}, in particular, we'll discuss our collision representations and collision-detection queries.

\gl scenes include a Fetch robot \cite{fetchrobot}, movable YCB objects \cite{calli2015ycb}, and 105 static (non-movable) ReplicaCAD scenes \cite{szot2021habitat}. Note that scenes in the ReplicaCAD dataset include some interactive furniture (e.g. openable cabinet drawers and doors), but we don't simulate these in \gl as they aren't required for the \reasy benchmark.

As discussed in \Cref{sec:system_abstracted_physics}, our approximate kinematic sim must perform collision queries between the robot (including grasped object, if any) and the environment (resting movable objects and the static scene). We represent each articulated link of the robot with a set of spheres (green in \Cref{fig:gala_collision_geometry}). These are authored manually, with the goal to approximate the shape of the Fetch robot with a minimal number of spheres. We also represent each grasped object with a set of spheres (blue). These are generated offline using a space-filling heuristic. Rather than supporting arbitrary sphere radius, we limit ourselves to a sphere-radius ``working set" of \{1.5 cm, 5 cm, 12 cm\}. This limitation is important as we'll see shortly.

We approximate a ReplicaCAD scene as a voxel-like structure called a column grid (gray in \Cref{fig:gala_collision_geometry}). A column grid is generated offline for a particular scene and a particular sphere radius from our working set, so we generate three column grids per scene. A column grid is a dense 2D array of columns in the XZ (ground) plane, with 3-centimeter spacing. For each column, we represent vertical free space as a list of layers. For example, a column in an open area of the room would contain just one layer, storing two floating-point height values roughly corresponding to the height of the floor and the height of the ceiling. A column in the vicinity of a table, meanwhile, would contain two layers: one spanning from the floor to the underside of the table, and another spanning from the table surface to the ceiling. Finally, the stored height values don't actually represent the surface heights themselves, but rather the height of the query sphere (of known radius) in contact with the surface. For ReplicaCAD scenes, the maximum number of layers for any column is approximately 10 and corresponds to columns in the vicinity of a particular bookshelf with many shelves. 

A column grid is generated offline using the ReplicaCAD scene's source triangle mesh and \hb's sphere-query functionality. We load the scene in \hb and use the scene extents to derive the column grid's XZ (ground-plane) extents. We iterate over this region using our chosen 3-cm spacing. For each column, we perform a brute-force search of the vertical region at the column's XZ position, using a series of sphere-overlap and vertical sphere-casts to find the free spans.

At runtime, to detect collisions between the robot (including grasped object, if any) and the static scene, we implement a fast sphere-versus-column-grid query. First, we select the appropriate column grid corresponding to the query sphere's radius. Second, we retrieve the nearest column corresponding to the sphere's XZ position. Finally, we linearly search the column's layers to determine whether the query sphere's Y position is in free versus obstructed space. This linear search is accelerated using caching: we start the search from the same layer index found in recent searches. This leverages spatial and temporal coherency, for example, consider the robot reaching under a table: if one sphere from the robot arm's link is found to be between the floor and the underside of a table, it's likely that other spheres from that same link or other queries from succeeding timesteps will also lie in that vertical layer.

Whereas a grasped movable object is represented with a set of spheres (blue in \Cref{fig:gala_collision_geometry}), a resting movable object is approximated as an oriented box (orange). This is computed from the YCB object's triangle mesh in a preprocess. At runtime, to detect collisions between the robot (including grasped object, if any) and the resting movable objects, we perform sphere-versus-box queries. There are generally 30 resting movable objects in the environment (1 target object and 29 distractor objects) and we need to avoid performing all 30 sphere-versus-box queries. So, we use a ``regular grid" acceleration structure to quickly retrieve a list of nearby resting objects.

Resting movable objects are inserted into a regular grid at episode initialization. This is a dense 2D array spanning the XZ (ground) plane, with each cell storing a list of objects that overlap it. Objects will generally overlap multiple cells and thus be present in the object lists of multiple cells. When an object is grasped by the robot, it is removed from all relevant cells in the regular grid, and if the object is later dropped, it is re-inserted into the regular grid at its new resting position.

Let's consider how to find the list of nearby resting objects for a given query sphere. The regular grid spacing is chosen such that cells are at least $4\times$ the largest radius in our sphere-radius working set (12 cm). A query sphere may overlap up to four adjacent cells in the regular grid, e.g. the sphere is centered near the shared edge of two cells or the shared corner of four cells. A naive approach here would be to merge and de-duplicate the object lists of the four cells. We avoid this expense and instead maintain \textit{four separate regular grids}, all spanning the entire scene XZ extent, with carefully-chosen varying X and Z offsets for the cell boundaries. In this way, any query sphere is guaranteed to lie fully inside a single cell of one of these grids (not spanning a cell edge or corner). Thus, our list of nearby resting objects is simply the list stored in this cell; we don't have to merge or de-duplicate multiple lists. Note this approach of four somewhat-redundant regular grids comes at the expense of extra memory and added insertion/removal compute time.

\section{Simulator Flexibility to new Assets}
\label{sec:new_assets} 
\begin{figure}[h!]
\centering
  \includegraphics[width=1.0\linewidth]{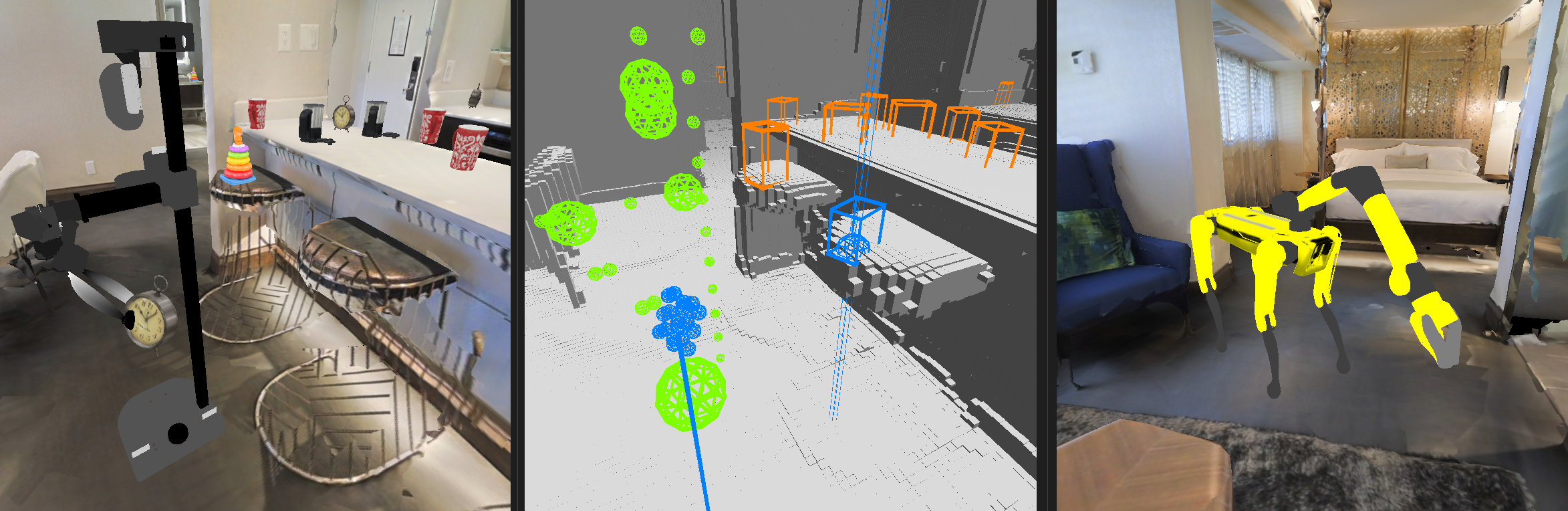}
  \caption{\gl with an MP3D scene, Google Scanned Objects \cite{downs2022google}, Stretch robot (left), Stretch debug viz (center), and Spot robot (right).}
  \label{fig:new_robots}
\end{figure}

\gl can work with various assets (robots, scenes and objects) from different sources. We use a mostly-automated pipeline that includes optimizing assets for the batch renderer and generating collision geometry (see \cref{sec:add_collision_detection_details}). In \cref{fig:new_robots} we added Stretch and Spot robots loaded in a scene from the MP3D dataset \cite{ramakrishnan2021habitat} with new objects.

\section{Description of Heuristics}
\label{sec:heuristics}

\subsection{Sliding Heuristic}
\label{sec:sliding_heuristics}

We implement the robot sliding heuristic as the following steps: (1) start from a candidate pose, (2) if the pose penetrates the scene, compute a jitter direction in the ground plane (3) jitter the robot base in the horizontal plane. Repeat from step 1 until a penetration-free  pose is found, up to 3 times. If this fails, the robot does not move on that step.

\subsection{Object Placing Heuristic}
\label{sec:placing_heuristics}

The object placement heuristic is as follows: (1) start from a candidate pose, (2) if the pose penetrates the scene, compute a jitter direction in the ground plane, with some randomness, (3) jitter the dropped object and re-cast down to a support surface. Repeat from step 1 until a penetration-free  pose is found, up to 6 times. If this fails, we restore the dropped object to its resting position prior to grasp. This approximates the dropped object bouncing or rolling away. Snap-to-surface is an instantaneous operation that resolves within one physics step; objects do not fall or settle over time.

\end{document}